\documentclass[letterpaper, 10 pt, conference]{ieeeconf}  

\IEEEoverridecommandlockouts                              
\usepackage{xcolor}
\usepackage[numbers]{natbib}
\let\labelindent\relax
\usepackage{enumitem}
\newlist{italicenum}{enumerate}{1}
\setlist[italicenum,1]{label=\textit{\alph*.}, font=\itshape}

\usepackage{amsmath} 
\usepackage{url}
\usepackage{graphicx}
\usepackage{hyperref}
 
\hypersetup{
    colorlinks=true,
    linkcolor=blue,
    filecolor=magenta,      
    citecolor=blue,
    urlcolor=blue,
}

\usepackage{algpseudocode}
\usepackage{algorithm}
\usepackage[algo2e]{algorithm2e}
\usepackage{multirow}
\usepackage[normalem]{ulem}
\usepackage{booktabs}
\usepackage{gensymb}
\usepackage{lipsum}
\usepackage{colortbl} 
\usepackage{amssymb}
\usepackage{pifont}

\usepackage{booktabs}       
\usepackage{amsfonts}       
\usepackage{nicefrac}       
\usepackage{arydshln}

\usepackage{amsmath,amsfonts,bm}
\usepackage{caption}
\usepackage{subcaption}

\def\eqref#1{equation~\ref{#1}}

\def\1{\bm{1}}

\DeclareMathAlphabet{\mathsfit}{\encodingdefault}{\sfdefault}{m}{sl}
\SetMathAlphabet{\mathsfit}{bold}{\encodingdefault}{\sfdefault}{bx}{n}

\definecolor{olivegreen}{HTML}{3C8031}
\newcommand{\xmark}{\ding{55}}%
\newcommand{\redcross}{\textcolor{red}{\xmark}}
\newcommand{\cmark}{\ding{51}}%
\newcommand{\greencheck}{\textcolor{olivegreen}{\cmark}}

\newcommand{\method}{\textsc{ViSk}}

\newcommand{\skin}{AnySkin}
\newcommand{\website}{\url{https://visuoskin.github.io/}}

\newcommand{\xxnote}[3]{}
\ifx\hidenotes\undefined
  \renewcommand{\xxnote}[3]{\color{#2}{#1: #3}}
\fi

\title{\LARGE \bf
Learning Precise, Contact-Rich Manipulation through \\Uncalibrated Tactile Skins
}

\author{
Venkatesh Pattabiraman$^{1,*}$ \quad
Yifeng Cao$^{2}$ \quad
Siddhant Haldar$^{1}$ \quad Lerrel Pinto$^{1}$ \quad 
Raunaq Bhirangi$^{1,3,*, \dagger}$ \thanks{$^{\dagger}$ Correspondence to: \texttt{raunaqbhirangi@nyu.edu}} \\[10pt]
$^{1}$ New York University \qquad $^{2}$ Columbia University \qquad $^{3}$ Carnegie Mellon University \\[10pt] * equal contribution \\[10pt]
\url{https://visuoskin.github.io/}
}

\begin{document}

\maketitle
\thispagestyle{empty}
\pagestyle{empty}

\begin{abstract}

While visuomotor policy learning has advanced robotic manipulation, precisely executing contact-rich tasks remains challenging due to the limitations of vision in reasoning about physical interactions. To address this, recent work has sought to integrate tactile sensing into policy learning. However, many existing approaches rely on optical tactile sensors that are either restricted to recognition tasks or require complex dimensionality reduction steps for policy learning. In this work, we explore learning policies with magnetic skin sensors, which are inherently low-dimensional, highly sensitive, and inexpensive to integrate with robotic platforms. To leverage these sensors effectively, we present the Visuo-Skin (\method{}) framework, a simple approach that uses a transformer-based policy and treats skin sensor data as additional tokens alongside visual information. Evaluated on four complex real-world tasks involving credit card swiping, plug insertion, USB insertion, and bookshelf retrieval, \method{} significantly outperforms both vision-only and optical tactile sensing based policies. Further analysis reveals that combining tactile and visual modalities enhances policy performance and spatial generalization, achieving an average improvement of 27.5\% across tasks.

\end{abstract}

\section{Introduction}
\label{sec:introduction}

Humans effortlessly perform precise manipulation tasks in their everyday lives, such as plugging in charger cords, or swiping credit cards -- activities that demand exact alignment and involve constrained motion. These tasks are so commonplace that we often overlook the complexity involved in executing them with the necessary accuracy. In contrast, much of the existing robot learning literature remains focused on simple, low-precision primitives such as pick-and-place, slide, push-pull, and lift that does not require such fine-grained spatial accuracy. As we strive to create robots capable of everyday tasks like handling cables and opening jars, it is crucial to develop frameworks that enable precise, contact-rich manipulation. 




\begin{figure}[tbp]
    \centering
    \includegraphics[width=\linewidth]{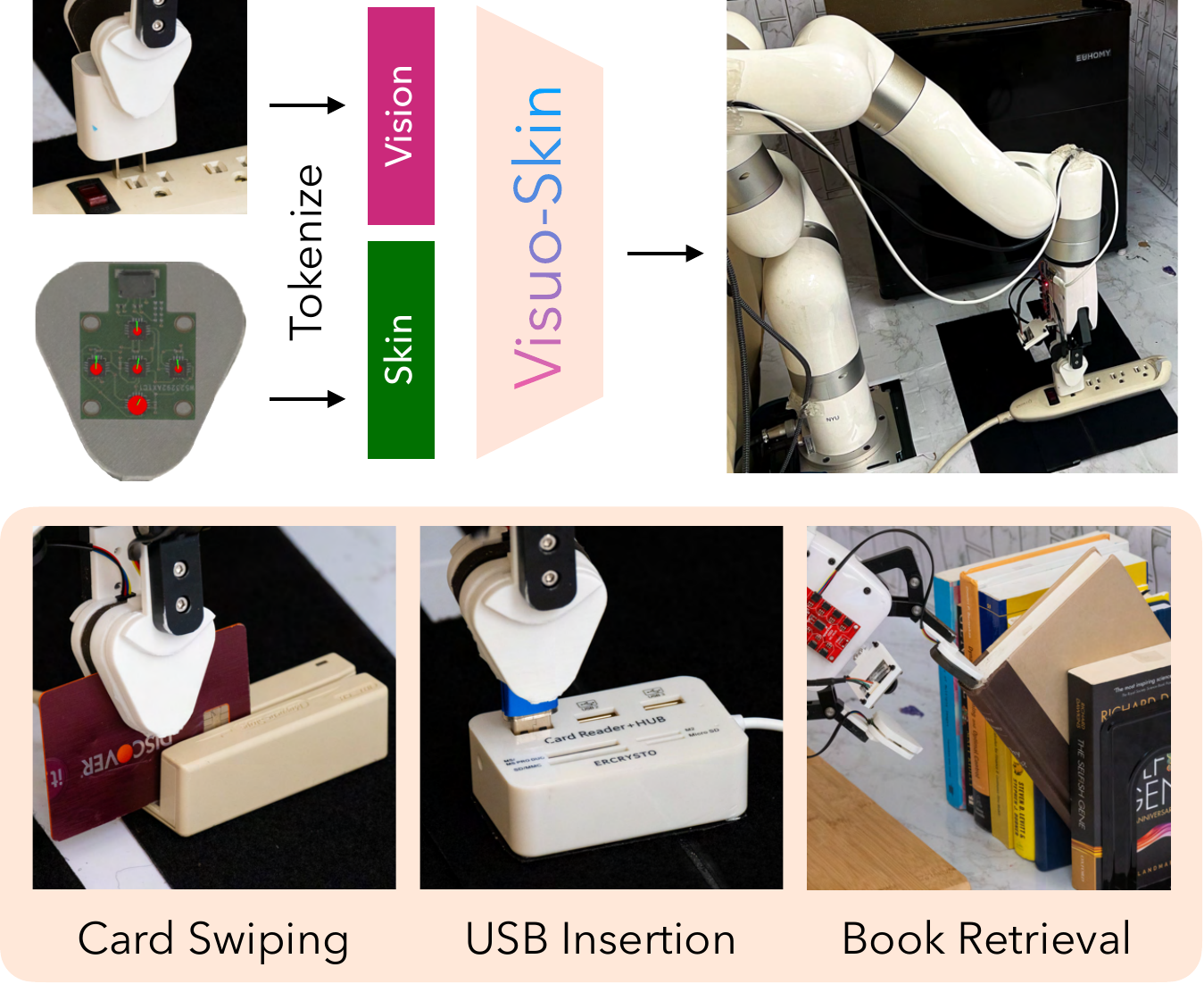}
    \caption{\method{} uses \skin{} with a simple transformer-based architecture to solve precise, contact-rich tasks.}
    \label{fig:enter-label}
\end{figure}

While the role of tactile feedback for robust execution of precise skills in humans is widely acknowledged~\cite{johansson1996sensory, jenner1982cutaneous}, analogous capabilities in robotic policies have lagged behind their vision-based counterparts. A variety of tactile sensors have been developed to bridge this gap in robotics, with optical tactile sensors like Gelsight~\cite{yuan2017gelsight} and DIGIT~\cite{lambeta2020digit} becoming popular choices in robot learning due to their high resolution. This increased resolution has facilitated several impressive works in areas like 3D reconstruction and localization~\cite{suresh2021tactile, suresh2023midastouch} and object recognition~\cite{funabashi2019morphology, bhirangi2023all}. However, the high dimensionality of tactile data from such sensors introduces additional complexity to the already challenging problem of policy learning. In most cases, the use of optical sensors necessitates dimensionality reduction through representation learning~\cite{lambeta2020digit}, explicit state estimation~\cite{li2014localization, kim2022active} or discretization~\cite{qi2023general, george2024visuo} to make it amenable to policy learning. This observation prompts an investigation into using alternative tactile sensing modalities that naturally offer lower-dimensional representations while still effectively capturing the essential characteristics of physical contact.

In this work, we present Visuo-Skin (\method{}), a simple framework for training precise robot policies using skin-based tactile sensing. \method{} uses a simple visuotactile policy architecture that incorporates tactile signals from \skin{}~\cite{bhirangi2024anyskin}, an affordable magnetic tactile sensor demonstrated to provide spatially continuous, low-dimensional (15-dimensional) sensing while being replaceable, making it well-suited for policy learning applications. The \method{} policy builds upon the BAKU~\cite{baku} architecture, which enables policy learning across multiple camera views and tasks. Through \method{}, we demonstrate that simply incorporating a tactile token obtained from a tactile encoder into state-of-the-art visual policy learning architectures enables effective visuotactile policy learning for precise real-world manipulation tasks that require visual as well as tactile inputs for localization. Furthermore, using a low-dimensional sensor like \skin{} allows policies to be learned end-to-end without requiring any task-specific preprocessing~\cite{li2014localization, kim2022active} of the tactile input or pretraining~\cite{lambeta2020digit, george2024visuo}. To the best of our knowledge, this work presents the first visuotactile framework enabling robots to perform precise contact-rich manipulation skills with policies that generalize across spatial variations while requiring a small number of robot demonstrations ($< 200$). 

To demonstrate the effectiveness of \method{}, we run extensive experiments on four precise manipulation tasks using a real-world xArm robot - \textit{plug insertion}, \textit{USB insertion}, \textit{credit card swiping} and \textit{bookshelf retrieval}. Our main findings are summarized below:

\begin{enumerate}[leftmargin=*,align=left]
    \item Policies trained with \method{} using skin-based tactile sensing exhibit an overall 27.5\% absolute improvement in performance compared to vision-only models across 4 precise manipulation tasks (Section ~\ref{subsec:visk-performance}).
    \item Through an ablation analysis, we study the impact of different modalities on policy learning, particularly the difference between visual and visuotactile policies for precise manipulation (Section ~\ref{subsec:visk-ablations}).
    \item Policies trained with the \skin{} tactile sensor~\cite{bhirangi2024anyskin} outperform those using optical tactile sensors such as DIGIT~\cite{lambeta2020digit} by at least  43\% on two real-world tasks, highlighting the benefits of skin-based sensors for visuotactile policy learning (Section ~\ref{subsec:digit-comparison}).
\end{enumerate}


All of our datasets, code for training, and robot evaluation will be made publicly available. Robot videos are best viewed at \website{}.

\section{Related Work}
\label{sec:related_work}

\subsection{Tactile sensing in Robotics}
Most robotic tasks involve physical interaction with the environment. Tactile sensing is critical in its ability to enable robots to reason about the physics of contact directly at the point of contact. Over the years, a number of diverse transduction mechanisms have been explored for tactile sensing. Resistive tactile sensors~\cite{sundaram2019learning, bhattacharjee2013tactile, stassi2014flexible} are inexpensive and relatively easy to fabricate, and provide discrete sensing making them well-suited for a range of applications that involve sensing the presence or absence of contact. Capacitative tactile sensors~\cite{glauser2019deformation, wu2020capacitivo} tend to provide more fine-grained measurements compared to resistive sensors and include proximity sensing in addtion to tactile sensing. Another versatile category of sensors are MEMS-based sensors~\cite{wettels2008biomimetic} that often combine multiple sensors such as audio and IMU sensors and can offer multimodal feedback in addition to higher resolution and mm-scale form factor.

Recently, optical tactile sensors like Gelsight~\cite{yuan2017gelsight} and DIGIT~\cite{lambeta2020digit} have emerged as a popular, high resolution alternative to existing tactile sensors for robotics due to a number of desirable properties such as their ease of replaceability and compatibility with well-understood neural architectures like convolutional neural networks~\cite{funabashi2019morphology}. Similarly, magnetic tactile sensors like Xela~\cite{tomo2018new} and ReSkin~\cite{bhirangi2021reskin} have garnered significant interest due to their scalable form factor, low dimensionality and ability to sense shear force in addition to their consistency across sensor instances~\cite{bhirangi2021reskin, suresh2023neural}. In light of these characteristics, the \method{} framework presented in this work uses \skin{}~\cite{bhirangi2024anyskin} a magnetic tactile sensor that strikes the right balance between low dimensionality and continuous contact sensing. Furthermore, its superior cross-instance signal consistency makes it more amenable than optical sensors to policy learning without the need for complex additional fabrication to prevent wear and tear~\cite{george2024visuo}.

\subsection{Visuotactile learning}
The meteoric rise of deep learning has paralleled recent developments in rapid prototyping and additive manufacturing. As a result, a number of recent works have investigated the use of machine learning for a host of tactile prediction tasks such as slip detection~\cite{li2018slip,9252140}, material classification~\cite{material_classification,baishya2016robust}, object identification~\cite{lin2019learning,schneider2009object} and 3D reconstruction~\cite{3drecon,ilonen2014three} across a range of tactile sensors. In this paper, we specifically focus on policy learning -- incorporating tactile information into robotic policies to enhance contact-rich manipulation. 

Recent works have demonstrated impressive improvements from incorporating tactile data into the policy learning framework for precise dexterity~\cite{guzey2023dexterity, guzey2024see} and bimanual manipulation~\cite{lin2024learning}. However, the high dimensional nature of dexterous control limits the task complexity and extent of generalizability enabled by these works. While \cite{qi2023general, yuan2024robot} use sim2real learning to demonstrate significant generalizability across objects for an in-hand rotation task, the task lacks precision, and sim2real transfer necessitates significant dilution of the tactile input to only capture coarse, discrete information. This limits the scalability of this approach to the precise, contact-rich tasks considered in this work. 

Yet other works rely on explicit pose estimation~\cite{kelestemur2022tactile} and handcrafted feature extraction~\cite{li2014localization, kim2022active} from optical tactile data for alignment when performing insertion tasks. While interesting, these techniques do not generalize to arbitrary tasks and require significant effort and domain knowledge to adapt to every new task. While some existing works have learned visuotactile policies for precise tasks such as insertion~\cite{lee2020making, li2022see}, all of these works evaluate performance in restricted settings with little to no spatial variation in the location of the insertion slot.  In this paper, we investigate visuotactile policy learning for contact-rich, high-precision tasks requiring spatial generalization, and conclusively show that \method{} policies use tactile feedback in conjunction with vision to substantially improve task performance.
\begin{figure*}[tbp]
    \includegraphics[width=\linewidth]{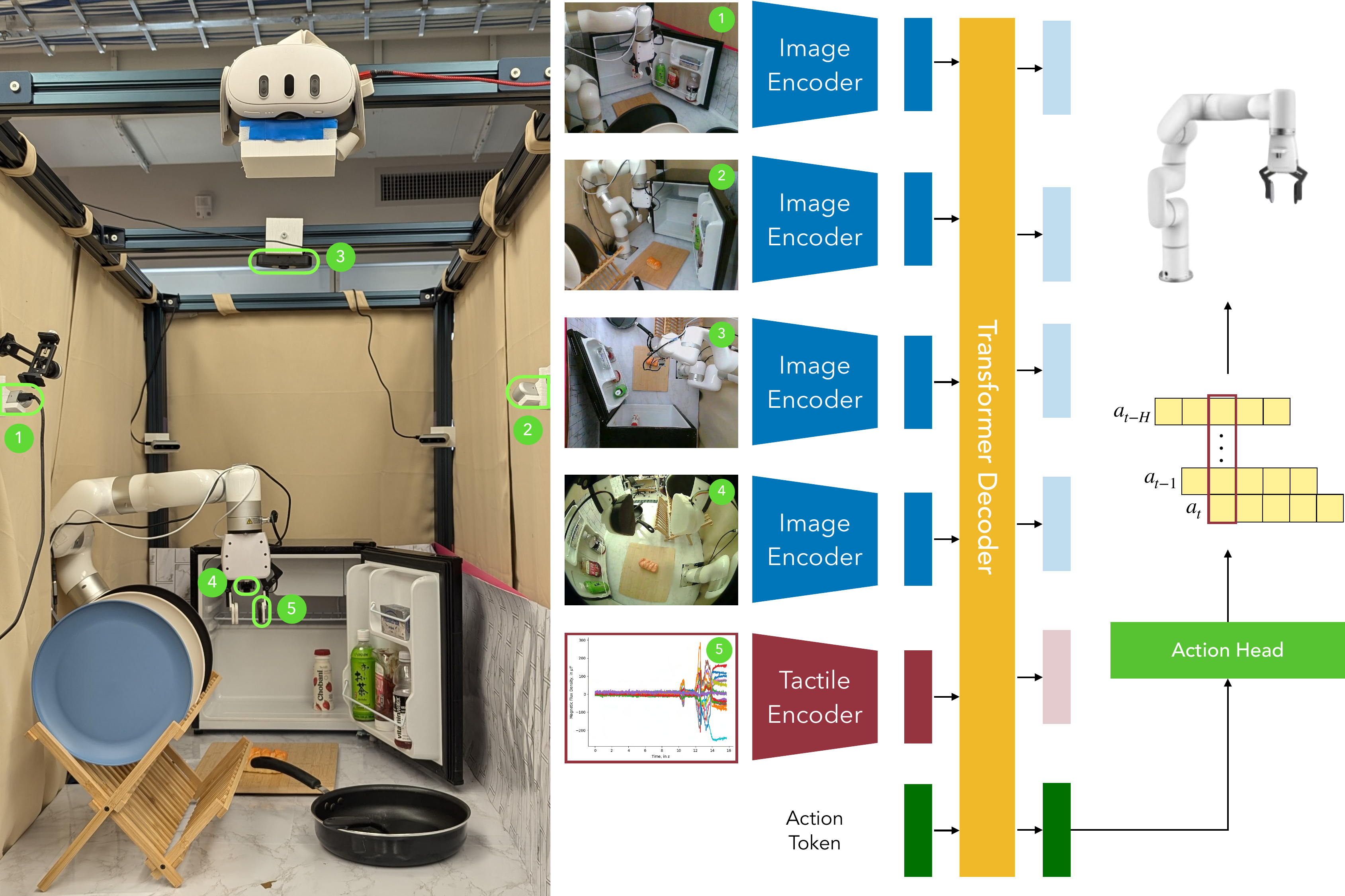}
    \caption{(left) Robot setup used for experiments in Section~\ref{sec:experiments}; (right) \method{} policy architecture uses ResNet-18~\cite{he2016deep} encoders for camera inputs and an MLP encoder for \skin{} input. An action token is appended to the encoded inputs before passing them through a transformer decoder, and the corresponding feature is used for action prediction by the action head.}
    \label{fig:expt-setup}
\end{figure*}

\section{Visuo-Skin Policy Learning (\method{})}
\label{sec:method}


Two key considerations in designing a framework for visuotactile policy learning include the choice of a tactile sensor capable of providing reliable tactile data across diverse environments and tasks, and designing a neural architecture able to effectively leverage multimodal visual and tactile information. Our proposed approach, \method{}, addresses these in the following ways: first, it employs AnySkin~\cite{bhirangi2024anyskin}, a magnetic tactile skin shown to yield consistent tactile measurements reliably under various conditions. Second, it builds upon state-of-the-art approaches to visual policy learning~\cite{baku} by incorporating a tactile encoding stream, allowing the network to effectively learn from multimodal data. Below, we describe each component of \method{} in detail.


\subsection{Data Collection}
\label{subsec:data_collection}
We use a VR-based teleoperation framework~\cite{openteach} employing the Meta Quest 3 headset to collect data for our real-world xArm robot experiments. Visual data from 4 camera views, including an egocentric camera attached to the robot gripper, is recorded at 30 Hz. Tactile data for the AnySkin experiments is recorded as magnetometer signals at 100 Hz, while data from the DIGIT sensors in comparative tests are recorded at 30 Hz, identical to the cameras. Drawing from prior work demonstrating the benefits of adding noise to demonstrations for policy learning~\cite{brandfonbrener2023visual, dasari2022rb2}, we add a uniformly sampled angular perturbation to the direction of the commanded robot velocity during teleoperation. This proves especially useful for increasing the diversity of contact-rich signals in the dataset by rendering the tasks slightly more challenging for the human operator. While large perturbations risk steering the learned behavior cloning policy astray, we find that injecting a minor directional noise yields an information-rich tactile signal while maintaining consistent task success.

\begin{figure*}[tbp]
    \centering
    \includegraphics[width=\linewidth]{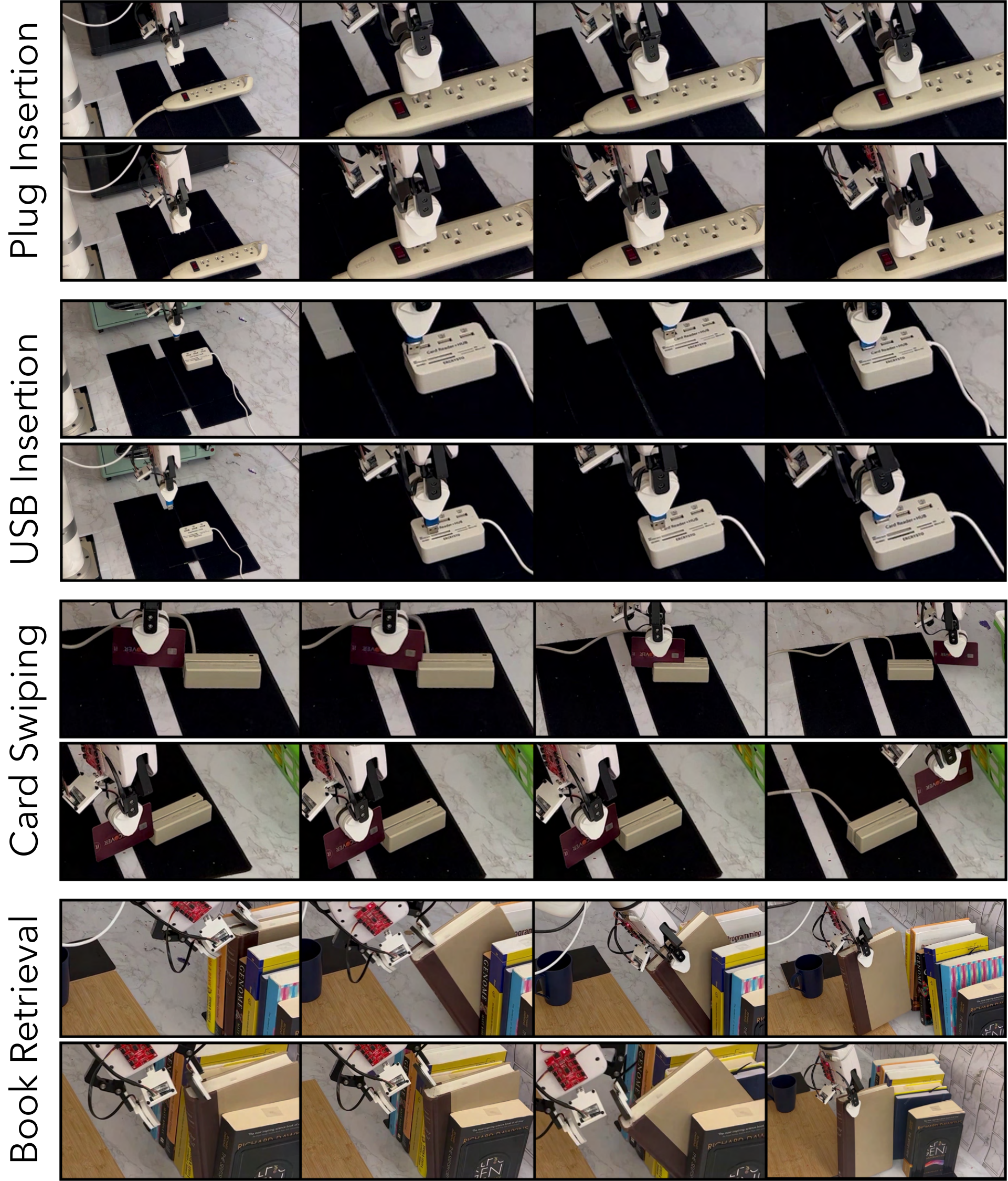}
    \caption{Close-up views of \method{} rollouts for the four tasks: Plug Insertion, Card Swiping, USB Insertion and Book Retrieval}
    \label{fig:rollouts}
    
\end{figure*}
    
\subsection{Policy Architecture}

The \method{} policy builds on top of BAKU~\cite{baku}, a state-of-the-art transformer-based policy learning architecture that learns visual policies across multiple camera views. Similar to BAKU, our architecture contains three main components:
\begin{figure*}[tbp]
    \centering
    \includegraphics[width=\linewidth]{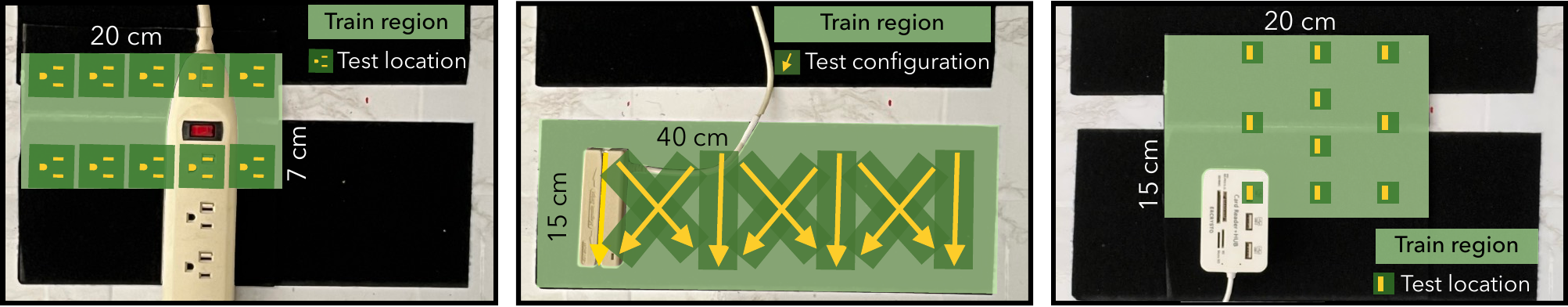}
    \caption{Overhead view depicting variations in target object locations for training and evaluation for plug insertion, card swiping and USB insertion (left to right). The enclosing light green box denotes the extent of variation in the training data. Test locations for plug insertion and USB insertion are marked on the image. For the card swiping task, arrows denote test locations and orientations of the card machine used for evaluation. For the book retrieval task (not depicted here), the order of books is randomized for every training demonstration, and test configurations consist of orderings unseen in training data.}
    \label{fig:variations}
\end{figure*}
\paragraph{Sensory Encoders} Visual inputs from cameras are encoded using a modified ResNet-18~\cite{he2016deep} visual encoder. Low-dimensional tactile inputs from the AnySkin sensor are encoded with a two-layer multilayer perceptron (MLP). Drawing from~\cite{bhirangi2021reskin}, we subtract a baseline measurement from each tactile reading to account for sensor drift. The encoded representations for each modality are projected to the same dimensionality to facilitate combining modalities in the observation trunk. Some of the ablations and comparisons presented in Section~\ref{sec:experiments} also use DIGIT sensors and robot proprioception as inputs to the policy. In line with prior works using DIGIT sensors for policy learning~\cite{lin2019learning, li2019connecting}, tactile images from the DIGIT sensor are encoded using the same ResNet-18 encoder as the visual data.  The proprioceptive inputs are encoded using a two-layer MLP.



\paragraph{Observation Trunk} The encoded inputs from all camera views, robot proprioception, and the tactile signals are treated as separate observation tokens and passed through a transformer decoder network~\cite{transformer}. A learnable action token is appended to the list of observation tokens and is used to obtain action features.

\paragraph{Action Head} Finally, an action head takes as input the action features from the observation trunk and predicts the corresponding actions. We found a deterministic action head learned using a mean squared error loss to suffice for our experiments. Considering the temporal correlation in robot movements, we follow prior work~\cite{baku,aloha,diffusionpolicy} and include action chunking to counteract the covariate shift often seen in the low-data imitation learning regime. During inference, we apply exponential temporal smoothing~\cite{aloha} for producing smoother robot motions. Our full policy architecture is depicted in Figure~\ref{fig:expt-setup}.







\section{Experiments}
\label{sec:experiments}
We study the effectiveness of the \method{} framework in a policy learning setting using behavior cloning. Our experiments are designed to answer the following questions:
\begin{itemize}[leftmargin=*,align=left]
    \item How does \method{} perform on precise manipulation tasks?
    \item How do different inputs affect performance of \method{}?
    \item Does \method{}'s use of \skin{} improve over DIGIT~\cite{lambeta2020digit}?
    \item Do \method{} policies generalize to unseen task variations?
\end{itemize}


\subsection{Environment Setup}

We use a Ufactory xArm 7 robot with its standard two-fingered gripper for all our experiments. To enable tactile sensing, we attach \skin{} sensor tips to the left gripper finger. An identically shaped, plain silicone tip is attached to the right finger. For baseline comparisons with the DIGIT sensor, we use a DIGIT sensor on either fingertip in line with prior work~\cite{li2018slip}. The camera inputs comprise synchronized RGB images at 128x128 resolution from three static third-person cameras and an egocentric camera mounted on the gripper. The action space is the change in the end-effector pose and gripper state. Our experimental setup is depicted in Figure~\ref{fig:expt-setup}. Learned policies are deployed at a 10Hz frequency. 


\subsection{Task Descriptions}
\label{sec:task-descriptions}
For all the analysis presented in this paper, we focus on a set of four contact-rich tasks that require high precision as well as spatial generalization. Each task has a target object that the robot must interact with, whose position is varied during demo collection. All evaluations use a fixed set of ten target locations unseen in the training demonstration data.

\paragraph{Plug Insertion} This task requires the robot to insert a plug into the first socket on a power strip. The arm starts with the plug grasped and the power strip randomly positioned within a 20cm $\times$ 7cm grid with a fixed orientation. The training dataset consists of 96 demonstrations.
\paragraph{USB Insertion} This task has the robot plugging a USB stick into a specific port on a USB hub. The arm starts with the USB stick grasped and the hub is positioned randomly within a 20cm $\times$ 15cm grid. The training dataset consists of 98 demonstrations.
\paragraph{Card Swiping} This task involves swiping a credit card through a card reader. The arm starts with the credit card grasped and the card reader randomly positioned within a 40cm $\times$ 15cm grid, and oriented at a random angle in the range $(-30^\circ, 30^\circ)$ from the direction the robot is facing. The training dataset consists of 90 demonstrations.
\paragraph{Book Retrieval} This task requires the robot to retrieve a specific book from a set of eight books placed together, with the order of books randomized each time. The robot must first reach for the target book, pivot it about its edge, and then grasp and pull it out of the bookrack. The training dataset consists of 172 demonstrations.

For the first three tasks, where the robot starts with a grasped object, we do not enforce hard constraints on the grasping location and allow some variability across runs. The extent of variation in target object configurations are shown in Fig.~\ref{fig:variations}. Evaluations are performed on a set of 10 held-out configurations for each task. 

\begin{table*}[tbp]
    \centering
    \caption{Success rates (out of 10) averaged over three seeds for policies trained on four tasks: Plug Insertion, USB Insertion, Card Swiping and Book Retrieval. \method{} policies are highlighted in grey.}
    \label{tab:performance}
    \begin{tabular}{ccccccccc}
        \toprule
         \multirow{3}{*}{Tactile Sensor} & \multicolumn{3}{c}{Input Modalities} & \multicolumn{4}{c}{Policy performance} \\ \cmidrule{2-9}
         & \begin{tabular}[c]{@{}c@{}}3rd Person\\ Camera\end{tabular} & \begin{tabular}[c]{@{}c@{}}Wrist\\ Cameras\end{tabular} & \begin{tabular}[c]{@{}c@{}}Robot\\ Proprioception\end{tabular} & Plug Insertion & USB Insertion & Card Swiping & Book Retrieval \\ \midrule
         \multirow{4}{*}{None} & \greencheck & \redcross   & \redcross     &$0.0 \pm 0.0$   & $0.7 \pm 0.6$ & $3.3 \pm 1.6$  &   $2.0 \pm 1.0$              \\
          & \greencheck & \redcross   & \greencheck &   $0.0 \pm 0.0$& $0.0 \pm 0.0$ &  $3.0 \pm 1.0$ & $0.6 \pm 0.5$                \\
          & \greencheck & \greencheck & \redcross     & $3.6 \pm 0.5$ & $2.3 \pm 2.0$ & $1.3 \pm 0.5$ & $3.3 \pm 1.1$                   \\
          & \greencheck & \greencheck & \greencheck & $1.0 \pm 1.0$ & $2.0 \pm 1.0$ & $3.0 \pm 1.7$ & $2.3 \pm 1.5$                   \\ \midrule
        \rowcolor[HTML]{EFEFEF} & 
          \greencheck & \redcross   & \redcross   & $2.3 \pm 1.1$ & $2.0 \pm 1.0$ & $\mathbf{7.0 \pm 1.7}$  & $3.6 \pm 2.5$                   \\
        \rowcolor[HTML]{EFEFEF} &  \greencheck & \redcross & \greencheck & $1.3 \pm 0.5$ & $1.0 \pm 1.0$ & $2.6 \pm 1.5$ &  $2.6 \pm 0.5$                  \\ 
        \rowcolor[HTML]{EFEFEF} & 
          \greencheck & \greencheck & \redcross & $\mathbf{6.6 \pm 1.5}$ & $\mathbf{5.6 \pm 1.5}$ & $1.0 \pm 1.0$ & $\mathbf{5.3 \pm 2.0}$                   \\ 
        \rowcolor[HTML]{EFEFEF} \multirow{-4}{*}{\skin{} (\method{})} & 
          \greencheck & \greencheck & \greencheck & $3.6 \pm 1.5$ & $2.0 \pm 1.0$ & $3.0 \pm 1.7$ & $4.6 \pm 2.0$                   \\  \midrule
         \multirow{2}{*}{DIGIT} & \greencheck & \redcross   & \redcross     & $2.3 \pm 0.5$  & $0.0 \pm 0.0$ & N/A  & N/A                \\
          & \greencheck & \greencheck & \redcross     & $1.6 \pm 1.5$ & $0.3 \pm 0.5$ & N/A & N/A                   \\\bottomrule
    \end{tabular}
\end{table*}
\subsection{Performance of \method{} policies}
\label{subsec:visk-performance}
We evaluate the performance of \method{} policies on the aforementioned precise manipulation tasks in the real world. To account for the high variance in performance of behavior cloning policies, we train policies across 3 random seeds and conduct 10 trials per seed for a total of 30 trials per evaluation. We report the aggregated success rate across seeds in Table~\ref{tab:performance}, and find that \method{} policies consistently outperform other variations across tasks.

Additionally, we observe that \method{} policies exhibit emergent seeking behavior. For instance, with the plug insertion and USB insertion tasks, we find that the policy first gets close to the location of the target (socket or port respectively), makes contact, and proceeds to move around as it tries to find the target. Once it seems to have located a change in contact characteristics, the policy pushes down and inserts successfully. This behavior is strong evidence of \method{} policies effectively leveraging tactile information from \skin{}. Further, it is distinctly different from the behavior of vision-only policies that simply attempt to push downwards once they get close to the insertion location regardless of alignment with the target. We see an analogous trend with the card swiping task, where the \method{} policy slows down as the card approaches the machine, and attempts alignment through contact before performing the swiping motion. The vision-only policy, on the other hand, seems to skip the alignment phase, and directly attempts to swipe the card, often entirely missing the card slot as a result. These failure modes demonstrates that purely visual policies lack the fine-grained tactile information that makes \method{} extremely effective on contact-rich, precise manipulation.

Similarly, for the book retrieval task, prominent failure modes for policies without \skin{} involve either applying too little force causing the book to flip back into the bookrack, or too much force causing the book to topple over entirely. \method{} policies apply a controlled downward force that enables them to pivot the book to an appropriate tilt, followed by grasping and retrieval as shown in Fig.~\ref{fig:rollouts}. Furthermore, for this task, repeated interaction with the sharp edges of the book caused the \skin{} to tear. All evaluations for this task reported in Table~\ref{tab:performance}, therefore, use a new instance of \skin{}. The sustained performance improvement of \method{} policies over vision-only policies even with replaced \skin{} is consistent with prior work~\cite{bhirangi2024anyskin} and underscores the importance of \skin{} to the \method{} framework.

\subsection{Effect of different input modalities on performance}
\label{subsec:visk-ablations}
From Table~\ref{tab:performance}, we find that while the addition of \skin{} inputs to the policy consistently improves performance, the addition of other modalities like the wrist camera and proprioception can have significant impact on policy performance depending on the task. A few consistent patterns emerge across tasks: (1) \method{} results in a significant improvement ($\geq 2\times$) in performance over the next best model, indicating its effectiveness on precise, contact-rich manipulation. (2) Adding proprioceptive input almost always results in a drop in performance. This can be attributed to the learned policy overfitting to proprioceptive information which is detrimental to tasks requiring spatial generalizability over target object locations. (3) With the exception of the card swiping task, the addition of a wrist camera improves policy performance. The wrist camera gives the policy a local visual understanding of the scene in the frame of the gripper, and in turn, the same frame as the robot's action space. This is especially useful for the more fine-grained adjustments required for high-precision tasks. For the card swiping task, visualization of demonstration data indicated that the wrist camera cannot see the card reader due to occlusion from the gripper and therefore simply acts as a noise input to the policy.

While the drops in performance due to proprioception as well as due to the wrist camera in the card swiping task could potentially be addressed by collecting more demonstrations, they highlight the true potential of the \method{} framework. The addition of \skin{} and the use of a transformer-based architecture enable the policy to incorporate reliable tactile feedback directly from the interface between the robot and the object being interacted with. The low dimensional nature of \skin{} signal eliminates the need for dimensionality reduction or intermediate representation learning and enables end-to-end learning of visuotactile policies from relatively few ($<200$) demonstrations.

\subsection{Comparison between \skin{} and DIGIT}
\label{subsec:digit-comparison}
To further highlight the role of \skin{} in the \method{} framework for precise manipulation tasks, we collect similar demonstration datasets for two of the tasks presented in Section~\ref{sec:task-descriptions} (Plug Insertion and USB Insertion) using DIGIT sensors instead of \skin{} sensors. We maintain the same policy architecture as \method{} with the exception of the tactile encoder, where we replace the MLP with a modified ResNet-18 architecture identical to the image encoders used for camera inputs. We train two variants of the DIGIT-based policies: one with raw DIGIT measurement as input to the policy, and another with the DIGIT measurement at the start of the trajectory subtracted from every subsequent measurement. We report statistics for the best-performing alternative. While the use of a different tactile sensor necessitates collection of new demonstration data, we try to keep the DIGIT and \skin{} datasets as close to each other as possible. Target object locations used for training as well as evaluation are identical between the experiments corresponding to both sensors. The results in Table~\ref{tab:performance} also compare the performance of \method{} using the skin-based \skin{} tactile sensor against the optical DIGIT~\cite{lambeta2020digit} sensor. 

We find that across both tasks, policies trained with \skin{} significantly outperform those trained with DIGIT. This difference could be attributed to the lower sensitivity of the DIGIT sensor making it difficult to detect small tactile signals from extrinsic contact of the grasped object. Furthermore, the significantly higher dimensionality of DIGIT observations compared to \method{} might also make it more difficult to learn a sensory encoder without overfitting to the training data. These experiments highlight the suitability of \skin{} over optical sensors for efficiently learning visuotactile policies for precise tasks, due to its ability to sense finer tactile details as well as its low dimensionality resulting in more robust policies. 




\subsection{Generalization to Unseen Task Variations}
To further probe the strengths of the \method{} framework, we investigate performance on unseen task variations for all of the tasks presented above. For each variation, we evaluate the best-performing \method{} policy for the respective task on the same set of target object configurations shown in Fig.~\ref{fig:variations} and present the results in Table~\ref{tab:variations-all}. Additionally, we also report generalization performance of a vision-only baseline, which is essentially the \method{} policy without tactile information.


\subsubsection{Plug Insertion}
\begin{figure}[tbp]
    \centering
    \includegraphics[width=0.99\linewidth]{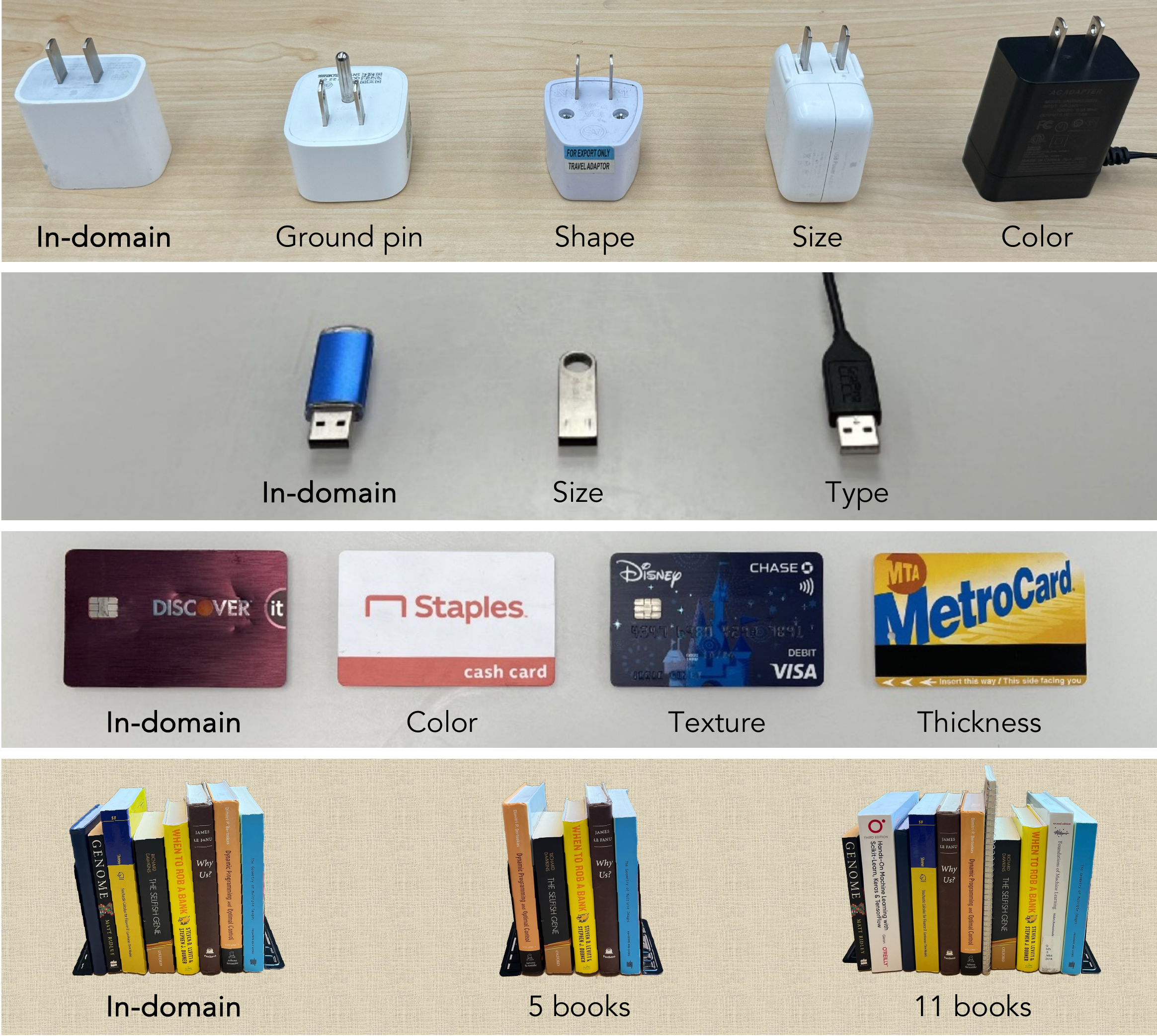}
    \caption{We vary different parameters of the object used for collecting demonstrations to analyze the generalizability of \method{} policies for the four tasks: (top to bottom) Plug Insertion, USB Insertion, Card Swiping and Book Retrieval.}
    \label{fig:generalization}
\end{figure}

We study the efficacy of the best-performing \method{} policy on four different variations of the plug as shown in Fig.~\ref{fig:generalization} -- addition of a ground pin, shape, size and color. On this small sample set, the \method{} policy generalizes surprisingly well to every plug variation except color despite their pins being in significantly different positions relative to the plug used for training. This is further evidence of \method{} policies effectively leveraging vision and touch even when faced with object variations distinctly different from training. Change in color is one variant where we see a significant drop in performance. The behaviors corresponding to these failures are qualitatively similar to the ablation without wrist cameras reported in Table~\ref{tab:performance}. This indicates that the policy might struggle to locate the socket when the wrist camera image is sufficiently out of distribution, further emphasizing the importance of wrist camera information in performing precise tasks like insertion.

\begin{table*}[htbp]
    \centering
    \small
    \caption{Performance of the best \method{} policy on different variations of each task. For the plug insertion, card swiping, and USB insertion tasks, we vary different parameters of the plug, card, and USB stick respectively. For the book retrieval task, we vary the number of books in the bookrack. We also report performance of a vision-only baseline policy for comparison.}
    \label{tab:variations-all}
    \footnotesize
    \setlength{\tabcolsep}{4pt}
    \begin{tabular}{lcccccc}
        \toprule
        \multirow{2}{*}{Task} & \multirow{2}{*}{Policy} & \multicolumn{5}{c}{Successful trials} \\\cmidrule{3-7}
         & & In-domain & \multicolumn{4}{c}{Variations}  \\\midrule
         \multirow{3}{*}{Plug Insertion} & & & Ground pin & Shape & Size & Color \\ 
         & ViSk & \textbf{8/10} & \textbf{6/10} & \textbf{6/10} & \textbf{6/10} & 1/10 \\
         & Vision-only & 5/10 & 1/10 & 2/10 & 4/10 & 1/10 \\
         \midrule
         \multirow{3}{*}{USB Insertion} & & & \multicolumn{2}{c}{Color/Type} & \multicolumn{2}{c}{Color/Size} \\
         & ViSk & \textbf{7/10} & \multicolumn{2}{c}{\textbf{5/10}} & \multicolumn{2}{c}{\textbf{4/10}} \\
         & Vision-only & 4/10 & \multicolumn{2}{c}{1/10} & \multicolumn{2}{c}{2/10}\\
         \midrule
         \multirow{3}{*}{Card Swiping} & & & Color & Color/Texture & Color/Thickness \\
         & ViSk & \textbf{9/10} & \textbf{8/10} & \textbf{6/10} & \textbf{7/10} & \\
         & Vision-only & 5/10 & 3/10 & 3/10 & 1/10 & \\
         \midrule
         \multirow{3}{*}{Book Retrieval} & & & \multicolumn{2}{c}{5 books} & \multicolumn{2}{c}{11 books} \\
         & ViSk & \textbf{7/10} & \multicolumn{2}{c}{\textbf{3/10}} & \multicolumn{2}{c}{\textbf{6/10}} \\
         & Vision-only & 4/10 & \multicolumn{2}{c}{2/10} & \multicolumn{2}{c}{4/10} \\
         \bottomrule         
    \end{tabular}
\end{table*}



\subsubsection{Card Swiping}
We similarly evaluate the performance of the best-performing \method{} policy on three different variations of the card as shown in Fig.~\ref{fig:generalization} -- color alone, color and texture (credit card with with embossed text on the surface), and color and thickness (paper-thin metrocard). As indicated by Table~\ref{tab:variations-all}, the \method{} policy generalizes surprisingly well to variations in color and thickness. This is further evidence of \method{} policies effectively leveraging vision and touch even when faced with object variations distinctly different from training. The relative performance drop due to variations in texture could be attributed to out-of-distribution tactile data resulting from stress concentrations at the locations of the embossed text.

\subsubsection{USB Insertion}
Similarly, for USB insertion, we study the effectiveness of the best-performing \method{} policy on two different variations of the USB stick as shown in Fig,~\ref{fig:generalization} -- color and type (USB cable), and color and size (different USB key). Results presented in Table~\ref{tab:variations-all} show that the performance of the \method{} policy drops by a small amount with the cable as well as the different USB stick. This drop can be attributed to the significant difference in both appearance as well as the surface properties of the objects used, and could potentially be bridged by increasing the number of training demonstrations and/or increasing the diversity of training data.
variations in shape, and also to a combination of change in color and addition of a cable. This is further evidence of \method{} policies effectively leveraging vision and touch even when faced with object variations substantially different from training.


\subsubsection{Book Retrieval}
Finally, we also evaluate the best-performing \method{} policy for variations of the book retrieval task, with different numbers of books in the bookrack as shown in Fig.~\ref{fig:generalization}. Our dataset is collected with 8 books, and we test generalizability to two variants with 5 and 11 books. For the 5 book variation, we start with the same initial arrangements as used in the original 10 evaluations, and randomly remove 3 books for each trial. For the 11-book variation, we randomize the order of the books for evaluation. Success rates are reported in Table~\ref{tab:variations-all}. We observe that despite prominent visual differences from the additional books, the \method{} policy is able to generalize well to the scenario with 11 books. This reinstates the effectiveness of the visuotactile representation learned in \method{} for generalizing to novel scenarios at inference. However, for the 5-book variation we find that performance drops significantly. Successful rollouts of the \method{} policy perform a pivoting motion as shown in Fig.~\ref{fig:rollouts} before grasping and retrieving the book. A qualitative analysis of the behaviors during failed rollouts seems to suggest that fewer books result in lower friction from the books neighboring the target book, precluding this pivoting motion. As a result, the target book either falls back into the bookrack or falls out onto the table.

Moreover, across the four tasks, we find that the vision-only baseline exhibits substantially worse generalization performance ie. shows larger drops in performance when different parameters of the object are varied. This indicates that \method{} leverages tactile feedback to improve robustness of learned policies to object variations, and highlights the value of tactile sensing to precise manipulation tasks.



\section{Conclusion and Limitation}
In this work, we presented Visuo-Skin (\method{}), a simple yet effective framework that leverages low-dimensional \skin{} tactile sensing for visuotactile policy learning in the real world. Our results demonstrate the efficacy of \method{} across a diverse range of precise, contact-rich manipulation tasks. Additionally, we also present a detailed analysis of the effect of different modalities on policy performance for this class of tasks and find that while the addition of wrist cameras can be critical to performance in tasks involving fine alignment, proprioception can often hurt spatial generalizability. We address a few limitations in this work: $(a)$ While \method{} shows significant improvements over vision-only policies, the policy's performance remains at approximately 60\% across all tasks. This suggests potential for further performance enhancement through fine-tuning the \method{} policy using reinforcement learning techniques~\cite{haldar2023teach}. $(b)$ Contrary to findings in prior studies, we observe that robot proprioception did not contribute to improved policy learning performance in precise manipulation tasks. This unexpected result warrants further investigation and presents an interesting direction for future research. $(c)$ All the tasks analyzed in this work involve maintained contact with the object throughout the duration of the task. Tasks that require making and breaking contact may involve specific nuances that could benefit from a similar detailed analysis. These limitations notwithstanding, we believe that \method{} presents a significant step in the right direction for advancing visuotactile policy learning in robotics.

\section*{Acknowledgments}
Special thanks for Krishna Bodduluri, Mike Lambeta, and team from Meta AI Research for providing the DIGIT sensors for comparison. Thanks to Tess Hellebrekers for providing the sensor skins and discussions for the experiments. This work was supported by grants from Honda, Hyundai, NSF award 2339096 and ONR awards N00014-21-1-2758 and N00014-22-1-2773. LP is supported by the Packard Fellowship.
\newpage






\bibliographystyle{IEEEtran}
\bibliography{IEEEabrv, references}

\begin{thebibliography}{10}
\providecommand{\url}[1]{#1}
\csname url@rmstyle\endcsname
\providecommand{\newblock}{\relax}
\providecommand{\bibinfo}[2]{#2}
\providecommand\BIBentrySTDinterwordspacing{\spaceskip=0pt\relax}
\providecommand\BIBentryALTinterwordstretchfactor{4}
\providecommand\BIBentryALTinterwordspacing{\spaceskip=\fontdimen2\font plus
\BIBentryALTinterwordstretchfactor\fontdimen3\font minus \fontdimen4\font\relax}
\providecommand\BIBforeignlanguage[2]{{%
\expandafter\ifx\csname l@#1\endcsname\relax
\typeout{** WARNING: IEEEtran.bst: No hyphenation pattern has been}%
\typeout{** loaded for the language `#1'. Using the pattern for}%
\typeout{** the default language instead.}%
\else
\language=\csname l@#1\endcsname
\fi
#2}}

\bibitem{johansson1996sensory}
R.~S. Johansson, ``Sensory control of dexterous manipulation in humans,'' in \emph{Hand and brain}.\hskip 1em plus 0.5em minus 0.4em\relax Elsevier, 1996, pp. 381--414.

\bibitem{jenner1982cutaneous}
J.~Jenner and J.~Stephens, ``Cutaneous reflex responses and their central nervous pathways studied in man,'' \emph{The Journal of physiology}, vol. 333, no.~1, pp. 405--419, 1982.

\bibitem{yuan2017gelsight}
W.~Yuan, S.~Dong, and E.~H. Adelson, ``Gelsight: High-resolution robot tactile sensors for estimating geometry and force,'' \emph{Sensors}, vol.~17, no.~12, p. 2762, 2017.

\bibitem{lambeta2020digit}
M.~Lambeta, P.-W. Chou, S.~Tian, B.~Yang, B.~Maloon, V.~R. Most, D.~Stroud, R.~Santos, A.~Byagowi, G.~Kammerer, \emph{et~al.}, ``Digit: A novel design for a low-cost compact high-resolution tactile sensor with application to in-hand manipulation,'' \emph{IEEE Robotics and Automation Letters}, vol.~5, no.~3, pp. 3838--3845, 2020.

\bibitem{suresh2021tactile}
S.~Suresh, M.~Bauza, K.-T. Yu, J.~G. Mangelson, A.~Rodriguez, and M.~Kaess, ``Tactile slam: Real-time inference of shape and pose from planar pushing,'' in \emph{2021 IEEE International Conference on Robotics and Automation (ICRA)}.\hskip 1em plus 0.5em minus 0.4em\relax IEEE, 2021, pp. 11\,322--11\,328.

\bibitem{suresh2023midastouch}
S.~Suresh, Z.~Si, S.~Anderson, M.~Kaess, and M.~Mukadam, ``Midastouch: Monte-carlo inference over distributions across sliding touch,'' in \emph{Conference on Robot Learning}.\hskip 1em plus 0.5em minus 0.4em\relax PMLR, 2023, pp. 319--331.

\bibitem{funabashi2019morphology}
S.~Funabashi, G.~Yan, A.~Geier, A.~Schmitz, T.~Ogata, and S.~Sugano, ``Morphology-specific convolutional neural networks for tactile object recognition with a multi-fingered hand,'' in \emph{2019 International Conference on Robotics and Automation (ICRA)}.\hskip 1em plus 0.5em minus 0.4em\relax IEEE, 2019, pp. 57--63.

\bibitem{bhirangi2023all}
R.~Bhirangi, A.~DeFranco, J.~Adkins, C.~Majidi, A.~Gupta, T.~Hellebrekers, and V.~Kumar, ``All the feels: A dexterous hand with large-area tactile sensing,'' \emph{IEEE Robotics and Automation Letters}, 2023.

\bibitem{li2014localization}
R.~Li, R.~Platt, W.~Yuan, A.~Ten~Pas, N.~Roscup, M.~A. Srinivasan, and E.~Adelson, ``Localization and manipulation of small parts using gelsight tactile sensing,'' in \emph{2014 IEEE/RSJ International Conference on Intelligent Robots and Systems}.\hskip 1em plus 0.5em minus 0.4em\relax IEEE, 2014, pp. 3988--3993.

\bibitem{kim2022active}
S.~Kim and A.~Rodriguez, ``Active extrinsic contact sensing: Application to general peg-in-hole insertion,'' in \emph{2022 International Conference on Robotics and Automation (ICRA)}.\hskip 1em plus 0.5em minus 0.4em\relax IEEE, 2022, pp. 10\,241--10\,247.

\bibitem{qi2023general}
H.~Qi, B.~Yi, S.~Suresh, M.~Lambeta, Y.~Ma, R.~Calandra, and J.~Malik, ``General in-hand object rotation with vision and touch,'' in \emph{Conference on Robot Learning}.\hskip 1em plus 0.5em minus 0.4em\relax PMLR, 2023, pp. 2549--2564.

\bibitem{george2024visuo}
A.~George, S.~Gano, P.~Katragadda, and A.~B. Farimani, ``Visuo-tactile pretraining for cable plugging,'' \emph{arXiv preprint arXiv:2403.11898}, 2024.

\bibitem{bhirangi2024anyskin}
R.~Bhirangi, V.~Pattabiraman, E.~Erciyes, Y.~Cao, T.~Hellebrekers, and L.~Pinto, ``Anyskin: Plug-and-play skin sensing for robotic touch,'' \emph{arXiv preprint arXiv:2409.08276}, 2024.

\bibitem{baku}
\BIBentryALTinterwordspacing
S.~Haldar, Z.~Peng, and L.~Pinto, ``Baku: An efficient transformer for multi-task policy learning,'' 2024. [Online]. Available: \url{https://arxiv.org/abs/2406.07539}
\BIBentrySTDinterwordspacing

\bibitem{sundaram2019learning}
S.~Sundaram, P.~Kellnhofer, Y.~Li, J.-Y. Zhu, A.~Torralba, and W.~Matusik, ``Learning the signatures of the human grasp using a scalable tactile glove,'' \emph{Nature}, vol. 569, no. 7758, pp. 698--702, 2019.

\bibitem{bhattacharjee2013tactile}
T.~Bhattacharjee, A.~Jain, S.~Vaish, M.~D. Killpack, and C.~C. Kemp, ``Tactile sensing over articulated joints with stretchable sensors,'' in \emph{2013 World Haptics Conference (WHC)}.\hskip 1em plus 0.5em minus 0.4em\relax IEEE, 2013, pp. 103--108.

\bibitem{stassi2014flexible}
S.~Stassi, V.~Cauda, G.~Canavese, and C.~F. Pirri, ``Flexible tactile sensing based on piezoresistive composites: A review,'' \emph{Sensors}, vol.~14, no.~3, pp. 5296--5332, 2014.

\bibitem{glauser2019deformation}
O.~Glauser, D.~Panozzo, O.~Hilliges, and O.~Sorkine-Hornung, ``Deformation capture via soft and stretchable sensor arrays,'' \emph{ACM Transactions on Graphics (TOG)}, vol.~38, no.~2, pp. 1--16, 2019.

\bibitem{wu2020capacitivo}
T.-Y. Wu, L.~Tan, Y.~Zhang, T.~Seyed, and X.-D. Yang, ``Capacitivo: Contact-based object recognition on interactive fabrics using capacitive sensing,'' in \emph{Proceedings of the 33rd annual acm symposium on user interface software and technology}, 2020, pp. 649--661.

\bibitem{wettels2008biomimetic}
N.~Wettels, V.~J. Santos, R.~S. Johansson, and G.~E. Loeb, ``Biomimetic tactile sensor array,'' \emph{Advanced robotics}, vol.~22, no.~8, pp. 829--849, 2008.

\bibitem{tomo2018new}
T.~P. Tomo, M.~Regoli, A.~Schmitz, L.~Natale, H.~Kristanto, S.~Somlor, L.~Jamone, G.~Metta, and S.~Sugano, ``A new silicone structure for uskin—a soft, distributed, digital 3-axis skin sensor and its integration on the humanoid robot icub,'' \emph{IEEE Robotics and Automation Letters}, vol.~3, no.~3, pp. 2584--2591, 2018.

\bibitem{bhirangi2021reskin}
R.~Bhirangi, T.~Hellebrekers, C.~Majidi, and A.~Gupta, ``Reskin: versatile, replaceable, lasting tactile skins,'' in \emph{5th Annual Conference on Robot Learning}, 2021.

\bibitem{suresh2023neural}
S.~Suresh, H.~Qi, T.~Wu, T.~Fan, L.~Pineda, M.~Lambeta, J.~Malik, M.~Kalakrishnan, R.~Calandra, M.~Kaess, \emph{et~al.}, ``Neural feels with neural fields: Visuo-tactile perception for in-hand manipulation,'' \emph{arXiv preprint arXiv:2312.13469}, 2023.

\bibitem{li2018slip}
J.~Li, S.~Dong, and E.~Adelson, ``Slip detection with combined tactile and visual information,'' in \emph{2018 IEEE International Conference on Robotics and Automation (ICRA)}.\hskip 1em plus 0.5em minus 0.4em\relax IEEE, 2018, pp. 7772--7777.

\bibitem{9252140}
J.~W. James and N.~F. Lepora, ``Slip detection for grasp stabilization with a multifingered tactile robot hand,'' \emph{IEEE Transactions on Robotics}, vol.~37, no.~2, pp. 506--519, 2021.

\bibitem{material_classification}
N.~Jamali and C.~Sammut, ``Majority voting: Material classification by tactile sensing using surface texture,'' \emph{IEEE Transactions on Robotics}, vol.~27, no.~3, pp. 508--521, 2011.

\bibitem{baishya2016robust}
S.~S. Baishya and B.~B{\"a}uml, ``Robust material classification with a tactile skin using deep learning,'' in \emph{2016 IEEE/RSJ International Conference on Intelligent Robots and Systems (IROS)}.\hskip 1em plus 0.5em minus 0.4em\relax IEEE, 2016, pp. 8--15.

\bibitem{lin2019learning}
J.~Lin, R.~Calandra, and S.~Levine, ``Learning to identify object instances by touch: Tactile recognition via multimodal matching,'' in \emph{2019 International Conference on Robotics and Automation (ICRA)}.\hskip 1em plus 0.5em minus 0.4em\relax IEEE, 2019, pp. 3644--3650.

\bibitem{schneider2009object}
A.~Schneider, J.~Sturm, C.~Stachniss, M.~Reisert, H.~Burkhardt, and W.~Burgard, ``Object identification with tactile sensors using bag-of-features,'' in \emph{2009 IEEE/RSJ International Conference on Intelligent Robots and Systems}.\hskip 1em plus 0.5em minus 0.4em\relax IEEE, 2009, pp. 243--248.

\bibitem{3drecon}
J.~Ilonen, J.~Bohg, and V.~Kyrki, ``Fusing visual and tactile sensing for 3-d object reconstruction while grasping,'' in \emph{2013 IEEE International Conference on Robotics and Automation}, 2013, pp. 3547--3554.

\bibitem{ilonen2014three}
------, ``Three-dimensional object reconstruction of symmetric objects by fusing visual and tactile sensing,'' \emph{The International Journal of Robotics Research}, vol.~33, no.~2, pp. 321--341, 2014.

\bibitem{guzey2023dexterity}
I.~Guzey, B.~Evans, S.~Chintala, and L.~Pinto, ``Dexterity from touch: Self-supervised pre-training of tactile representations with robotic play,'' \emph{arXiv preprint arXiv:2303.12076}, 2023.

\bibitem{guzey2024see}
I.~Guzey, Y.~Dai, B.~Evans, S.~Chintala, and L.~Pinto, ``See to touch: Learning tactile dexterity through visual incentives,'' in \emph{2024 IEEE International Conference on Robotics and Automation (ICRA)}.\hskip 1em plus 0.5em minus 0.4em\relax IEEE, 2024, pp. 13\,825--13\,832.

\bibitem{lin2024learning}
T.~Lin, Y.~Zhang, Q.~Li, H.~Qi, B.~Yi, S.~Levine, and J.~Malik, ``Learning visuotactile skills with two multifingered hands,'' \emph{arXiv preprint arXiv:2404.16823}, 2024.

\bibitem{yuan2024robot}
Y.~Yuan, H.~Che, Y.~Qin, B.~Huang, Z.-H. Yin, K.-W. Lee, Y.~Wu, S.-C. Lim, and X.~Wang, ``Robot synesthesia: In-hand manipulation with visuotactile sensing,'' in \emph{2024 IEEE International Conference on Robotics and Automation (ICRA)}.\hskip 1em plus 0.5em minus 0.4em\relax IEEE, 2024, pp. 6558--6565.

\bibitem{kelestemur2022tactile}
T.~Kelestemur, R.~Platt, and T.~Padir, ``Tactile pose estimation and policy learning for unknown object manipulation,'' \emph{arXiv preprint arXiv:2203.10685}, 2022.

\bibitem{lee2020making}
M.~A. Lee, Y.~Zhu, P.~Zachares, M.~Tan, K.~Srinivasan, S.~Savarese, L.~Fei-Fei, A.~Garg, and J.~Bohg, ``Making sense of vision and touch: Learning multimodal representations for contact-rich tasks,'' \emph{IEEE Transactions on Robotics}, vol.~36, no.~3, pp. 582--596, 2020.

\bibitem{li2022see}
H.~Li, Y.~Zhang, J.~Zhu, S.~Wang, M.~A. Lee, H.~Xu, E.~Adelson, L.~Fei-Fei, R.~Gao, and J.~Wu, ``See, hear, and feel: Smart sensory fusion for robotic manipulation,'' \emph{arXiv preprint arXiv:2212.03858}, 2022.

\bibitem{he2016deep}
K.~He, X.~Zhang, S.~Ren, and J.~Sun, ``Deep residual learning for image recognition,'' in \emph{Proceedings of the IEEE conference on computer vision and pattern recognition}, 2016, pp. 770--778.

\bibitem{openteach}
A.~Iyer, Z.~Peng, Y.~Dai, I.~Guzey, S.~Haldar, S.~Chintala, and L.~Pinto, ``Open teach: A versatile teleoperation system for robotic manipulation,'' \emph{arXiv preprint arXiv:2403.07870}, 2024.

\bibitem{brandfonbrener2023visual}
D.~Brandfonbrener, S.~Tu, A.~Singh, S.~Welker, C.~Boodoo, N.~Matni, and J.~Varley, ``Visual backtracking teleoperation: A data collection protocol for offline image-based reinforcement learning,'' in \emph{2023 IEEE International Conference on Robotics and Automation (ICRA)}.\hskip 1em plus 0.5em minus 0.4em\relax IEEE, 2023, pp. 11\,336--11\,342.

\bibitem{dasari2022rb2}
S.~Dasari, J.~Wang, J.~Hong, S.~Bahl, Y.~Lin, A.~Wang, A.~Thankaraj, K.~Chahal, B.~Calli, S.~Gupta, \emph{et~al.}, ``Rb2: Robotic manipulation benchmarking with a twist,'' \emph{arXiv preprint arXiv:2203.08098}, 2022.

\bibitem{li2019connecting}
Y.~Li, J.-Y. Zhu, R.~Tedrake, and A.~Torralba, ``Connecting touch and vision via cross-modal prediction,'' in \emph{Proceedings of the IEEE/CVF Conference on Computer Vision and Pattern Recognition}, 2019, pp. 10\,609--10\,618.

\bibitem{transformer}
A.~Vaswani, N.~Shazeer, N.~Parmar, J.~Uszkoreit, L.~Jones, A.~N. Gomez, {\L}.~Kaiser, and I.~Polosukhin, ``Attention is all you need,'' \emph{Advances in neural information processing systems}, vol.~30, 2017.

\bibitem{aloha}
T.~Z. Zhao, V.~Kumar, S.~Levine, and C.~Finn, ``Learning fine-grained bimanual manipulation with low-cost hardware,'' \emph{arXiv preprint arXiv:2304.13705}, 2023.

\bibitem{diffusionpolicy}
C.~Chi, S.~Feng, Y.~Du, Z.~Xu, E.~Cousineau, B.~Burchfiel, and S.~Song, ``Diffusion policy: Visuomotor policy learning via action diffusion,'' in \emph{Proceedings of Robotics: Science and Systems (RSS)}, 2023.

\bibitem{haldar2023teach}
S.~Haldar, J.~Pari, A.~Rai, and L.~Pinto, ``Teach a robot to fish: Versatile imitation from one minute of demonstrations,'' \emph{arXiv preprint arXiv:2303.01497}, 2023.

\end{thebibliography}

\end{document}